\documentclass[10pt,twocolumn,letterpaper]{article}
\usepackage[accsupp]{axessibility}  
\usepackage{multirow}
\usepackage{colortbl}
\usepackage{float}
\usepackage{comment}
\usepackage{pifont}
 \usepackage[pagenumbers]{cvpr} 

\definecolor{cvprblue}{rgb}{0.21,0.49,0.74}
\usepackage[pagebackref,breaklinks,colorlinks,allcolors=cvprblue]{hyperref}

\def\paperID{9921} 
\def\confName{CVPR}
\def\confYear{2025}

\title{DropoutGS: Dropping Out Gaussians for Better Sparse-view Rendering}

\author{
	Yexing Xu$^{1*}$ \quad Longguang Wang$^{1*}$ \quad Minglin Chen$^{1}$ \quad Sheng Ao$^{2}$ \quad Li Li$^{3}$ \quad Yulan Guo$^{1\dagger}$ \\
	\vspace{-0.8em} \\
	{ $^1$The Shenzhen Campus, Sun Yat-Sen University \quad $^2$Xiamen University \quad $^3$University of Macau}\\
	{\tt\small xuyx55@mail2.sysu.edu.cn, wanglg9@mail.sysu.edu.cn, guoyulan@sysu.edu.cn}\\
}

\begin{document}
\maketitle
\renewcommand{\thefootnote}{\fnsymbol{footnote}}
\footnotetext[1]{Equal contribution.}
\footnotetext[2]{Corresponding author.}
\begin{abstract}
Although 3D Gaussian Splatting (3DGS) has demonstrated promising results in novel view synthesis, its performance degrades dramatically with sparse inputs and generates undesirable artifacts. As the number of training views decreases, the novel view synthesis task degrades to a highly under-determined problem such that existing methods suffer from the notorious overfitting issue. 
Interestingly, we observe that models with fewer Gaussian primitives exhibit less overfitting under sparse inputs. 
Inspired by this observation, we propose a Random Dropout Regularization (RDR) to exploit the advantages of low-complexity models to alleviate overfitting. In addition, to remedy the lack of high-frequency details for these models, an Edge-guided Splitting Strategy (ESS) is developed. With these two techniques, our method (termed DropoutGS) provides a simple yet effective plug-in approach to improve the generalization performance of existing 3DGS methods.
Extensive experiments show that our DropoutGS produces state-of-the-art performance under sparse views on benchmark datasets including Blender, LLFF, and DTU. The project page is at:~\url {https://xuyx55.github.io/DropoutGS/}.
\end{abstract}    
\section{Introduction}
\label{sec:intro}
The computer vision community has witnessed incredible advances of Novel View Synthesis (NVS), which aims at synthesizing novel views of a scene from a set of observed views. Traditional methods employ explicit scene representations, such as point clouds \cite{fan2017point}, voxels \cite{choy20163d}, and meshes \cite{wang2018pixel2mesh}, to represent the 3D scene for novel view synthesis. Recently, learning-based methods have achieved remarkable progress compared to previous ones. Specifically, Neural Radiance Field (NeRF) \cite{NeRF} represents a scene with a neural network, achieving high-quality rendering and lightweight storage. 3D Gaussian Splatting (3DGS) \cite{3DGS} decomposes a scene into a set of discrete Gaussian primitives and utilizes a splat-based rendering technique to generate novel views in real time. Nevertheless, these methods require a large number of training views and suffer from notable performance drops with insufficient views, which limits their applications in real-world scenarios. 

To achieve novel view synthesis under sparse views, DRGS \cite{DRGS} encourages Gaussian primitives to align with the object surface by using the depth information obtained from a pre-trained monocular depth estimator. Then, DNGaussian \cite{DNGS} improves the depth regularization by further considering the smoothness of the predicted depth map. In addition, DNGaussian proposes a global-local depth normalization to address the scale inconsistency in monocular depth estimation. However, these methods are sensitive to the accuracy of the depth map and the depth error may be amplified to produce unsatisfactory artifacts.

\begin{figure}[t]
    \centering
    \includegraphics[width=\linewidth]{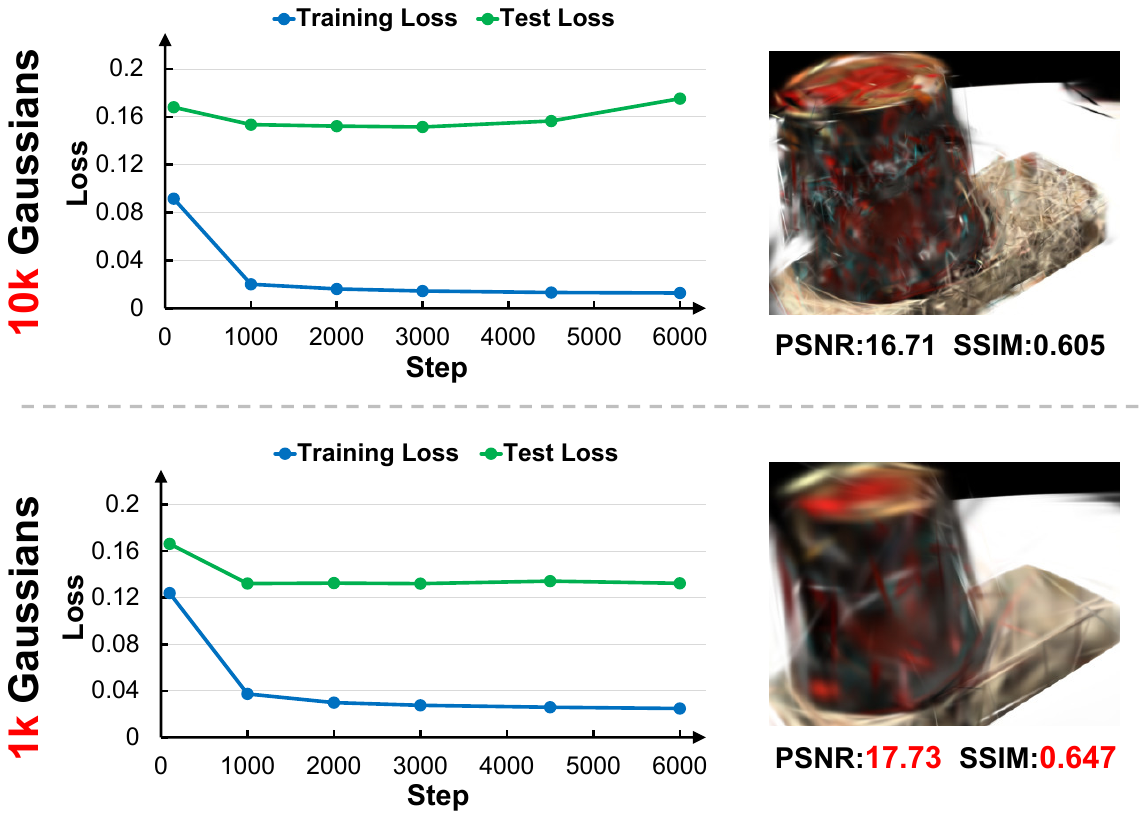}
    \caption{\textbf{Results produced by 3DGS with different numbers of Gaussians.} The training loss curves and rendered results are visualized for comparison. Compared with the model with 1k Gaussians, the one with 10k Gaussians suffers overfitting and produces inferior results.
    }
    \label{img:overfit}
\end{figure}

In this paper, we take a step to investigate the performance degradation of 3DGS under sparse views. Interestingly, we observe a notorious overfitting issue during the training of 3DGS, as illustrated in Fig.~\ref{img:overfit}. While the training loss of the model with 10k Gaussians continues to decrease, the test loss begins to increase after 3000 iterations. Meanwhile, the rendered image suffers severe hollow artifacts. In contrast, this phenomenon is weakened when the number of Gaussians is reduced to 1k and the rendered images are smoother. Nevertheless, the edge and texture become blurry. In summary, under sparse views, 3DGS is easily over-parameterized and prone to overfitting the training data without capturing the 3D structure. 

Inspired by this observation, we propose DropoutGS that combines an effective yet efficient dropout technique with 3DGS to address the overfitting issue under sparse views. Specifically, our DropoutGS randomly dropouts several Gaussians during training to alleviate overfitting. To remedy the detail loss caused by the dropout technique, we further introduce an Edge-guided Splitting Strategy (ESS) to encourage the Gaussians to focus more on edge regions. While previous sparse view synthesis methods commonly rely on exploring additional constraints (e.g., depth maps) to produce satisfactory performance, our DropoutGS attempts to handle this issue from another perspective of overfitting. Extensive experiments show that our DropoutGS produces competitive performance against previous methods on diverse benchmark datasets.

Overall, the main contributions are summarized as follows:

\begin{itemize}
    \item We study the performance degradation of 3DGS under sparse views and attribute it to the overfitting issue. From this point of view, we propose DropoutGS to leverage the dropout technique to alleviate this problem.

    \item We propose a Random Dropout Regularization (RDR) to alleviate overfitting and obtain smooth rendering results by randomly dropping out Gaussians during optimization.

    \item We introduce an Edge-guided Splitting Strategy (ESS) to encourage 3DGS to focus more on edge regions during optimization for finer details.
    
    \item Our DropoutGS is compatible with diverse 3DGS-based methods and achieves state-of-the-art performance on multiple benchmark datasets. 
\end{itemize}
\section{Related Work}
\label{sec:relatedwork}

\subsection{Novel View Synthesis}
Novel View Synthesis (NVS) is a long-standing task aiming at generating novel views of a scene from a set of observed images. In this field, radiance fields are widely used and produce remarkable progress. Neural Radiance Fields (NeRF) \cite{NeRF} encodes scenes into a neural network and achieves high-fidelity view synthesis by volumetric rendering. However, the computationally intensive rendering processes limit its scalability and efficiency. Later, a number of methods are developed to improve the rendering speed \cite{F2NeRF, SteerNeRF, FastNeRF, InstantNGP} and image quality \cite{MipNeRF, ZipNeRF, SurfelNeRF} of NeRF. More recently, the unstructured radiance field 3DGS proposed by Bernhard \textit{et al.} has achieved significant breakthroughs in novel view synthesis tasks. 3DGS represents complex scenes using a series of discrete Gaussian basis functions. In addition, efficient and differentiable splatting techniques are adopted for real-time rendering. Inspired by the success of 3DGS, subsequent methods further improve its representational capabilities \cite{2DGS, GES} and splitting strategies \cite{AbsGS, PixelGS, Minisplat}. Despite remarkable results, these methods suffer severe performance drops under sparse views.

To achieve novel view synthesis under sparse views, early works have explored various regularizations for NeRF, including depth \cite{DSNeRF, DDPNeRF, SparseNeRF}, normal \cite{FlipNeRF}, frequency \cite{FreeNeRF} and cross-view consistency \cite{GecoNeRF, Sparf}. As for recent 3DGS, depth prior \cite{DRGS, DNGS, FSGS, SparseGS} has also been widely studied to stabilize the optimization of Gaussians under sparse views. Despite remarkable progress, these methods still suffer two limitations. \textit{First}, these methods rely on an additional monocular depth estimation module to obtain the depth map, which introduces extra computational overhead. \textit{Second}, as 3DGS is sensitive to depth accuracy, depth errors could be accumulated and amplified to produce undesirable artifacts in the rendered images. Different from the aforementioned methods, we propose an alternative to consider the performance degradation as an overfitting problem and introduce a DropoutGS method.
 
\subsection{Overfitting in Deep Learning}
Overfitting is a notorious problem in deep learning. It usually arises from limited training data, high model complexity, or insufficient regularization, leading to poor generalization on unseen samples. To address this issue, various strategies have been studied. Specifically, dropout \cite{Dropout, Dropconnect, Bayesiandropout, Pooldropout}, as one of the most effective techniques, improves the robustness of the model by randomly deactivating a subset of neurons during training. Data augmentation strategies  \cite{cutout, erasing, mixup, cutmix} have been widely investigated to increase the diversity of training data. Adversarial training \cite{rice2020overfitting, wong2020fast, goodfellow2014explaining, madry2017towards} aims to introduce multiple perturbations like noise in the training process, encouraging the model to become more resilient and robust to outliers. 

In this paper, we observe a similar overfitting issue for 3DGS under sparse training views. Motivated by the great success of dropout to address this issue, we develop DropoutGS by incorporating a Random Dropout Regularization (RDR) to {smooth out the overfitting degradation} and an Edge-guided Splitting Strategy (ESS) to compensate for the detail loss.
\section{Method}
\begin{figure}[ht]
    \centering
    \includegraphics[width=1\linewidth]{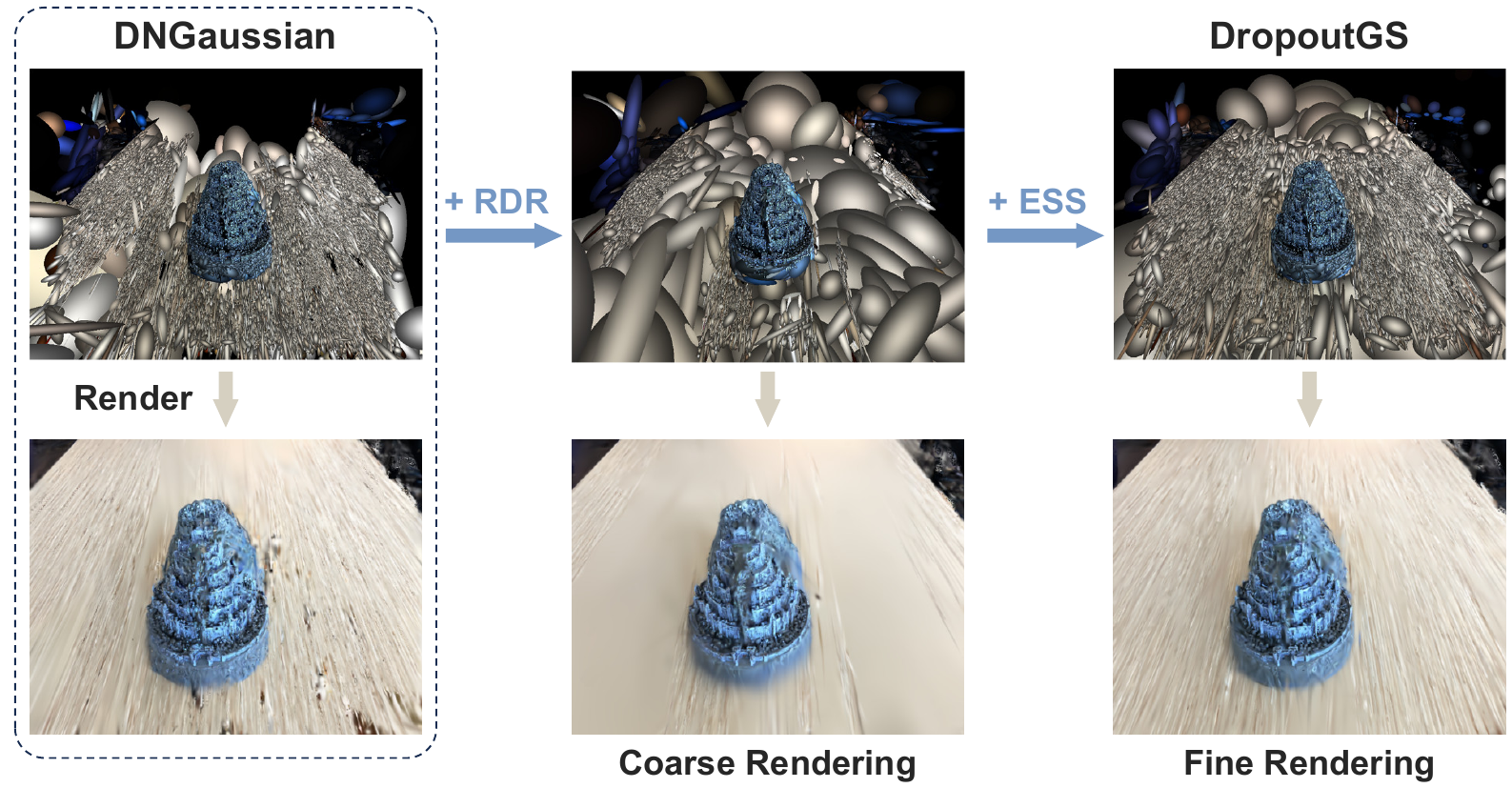}
    \caption{\textbf{An overview of our framework.} RDR is first employed to alleviate the overfitting issue. Then, ESS is adopted to split the large Gaussians to better capture high-frequency details.}
    \label{img:coarse-to-fine}
\end{figure}

Our DroupGS consists of a Random Dropout Regularization (RDR) and Edge-guided Splitting Strategy (ESS), as illustrated in Fig.~\ref{img:coarse-to-fine}.
In this section, we first present the preliminaries of 3DGS in Sec.~\ref{sec:Preliminaries}. Then, we conduct pilot experiments to illustrate our motivation in Sec.~\ref{sec:Motivation}. Next, we detail our method designs separately. Finally, we present the loss function of our method in Sec.~\ref{sec:loss}. 

\subsection{Preliminaries}
\label{sec:Preliminaries}
3DGS~\cite{3DGS} represents a 3D scene with a set of Gaussian primitives, each with trainable parameters including mean position $\mu$, 3D covariance $\Sigma$, scaling factor $s$, opacity $o$, and spherical harmonic features $f$. 3DGS employs the Gaussian basis functions to capture the spatial distribution of scene features. The rendering process blenders these basis functions to generate views, calculating the color as a weighted summation of Gaussian primitives: 
\begin{equation}
    C(x) = \sum_{i\in\mathcal{N}} \alpha_i \prod_{j=1}^{i-1}(1-\alpha_j)c_i = \sum_{i\in\mathcal{N}}w_{i}c_i,
    \label{eq:rendering}
\end{equation}
where $c_i$ is the decoded spherical harmonic features, and $w_i$ measures the contribution of Gaussians to the rendering results. 

\subsection{Pilot Study}
\label{sec:Motivation}
It is widely known that the imbalance between model complexity and training data amount is a primary reason for overfitting. For 3DGS, the number of Gaussian primitives determines the complexity of the model while the number of training views refers to the training data amount. From this point of view, we attribute the performance degradation of 3DGS under sparse views to the overfitting caused by an excessive number of Gaussian primitives.

\begin{figure}[t]
    \centering
    \includegraphics[width=1\linewidth]{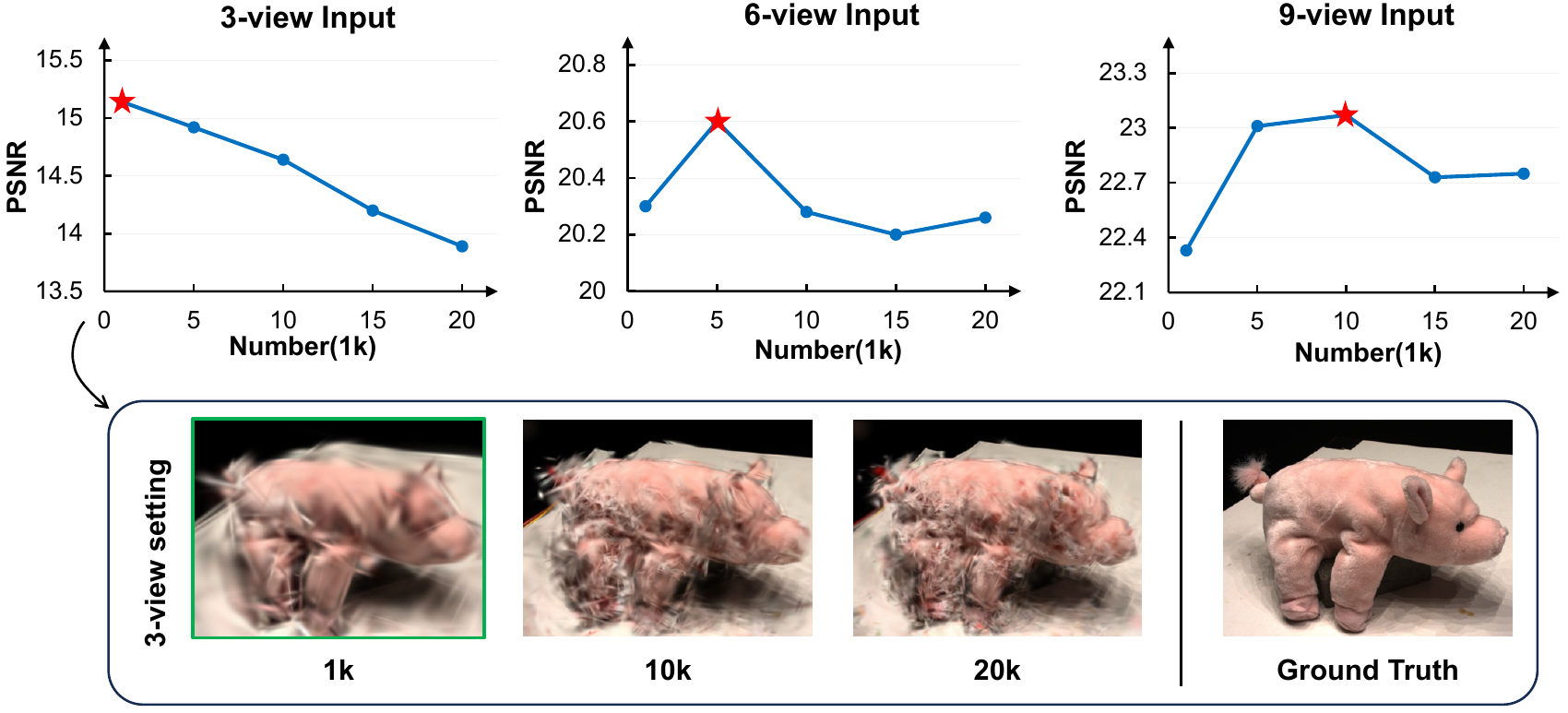}
    \caption{\textbf{The relationship between the amount of training data and model complexity.} We investigate the performance of 3DGS with different primitive settings under varying numbers of sparse views. The results show that models achieve optimal performance when their complexity matches the training data.}
    \label{img:motivation}
\end{figure}

To demonstrate this, we first investigate the performance of 3DGS with different numbers of Gaussian primitives under sparse training views. Specifically, we initialize the 3DGS using point clouds with different numbers of points and train these models on the DTU dataset. Note that, densification and pruning are not performed to maintain consistent model complexity throughout the training phase. As illustrated in Fig.~\ref{img:motivation}, the model with 1k primitives achieves the best performance under 3-view input while models with higher complexity suffer from performance degradation. Meanwhile, as the number of input views increases, the optimal model size grows from 1k to 10k. These observations demonstrate that an excessive number of Gaussian primitives leads to the overfitting issue and inferior performance.

\begin{figure}[t]
    \centering
    \includegraphics[width=\linewidth]{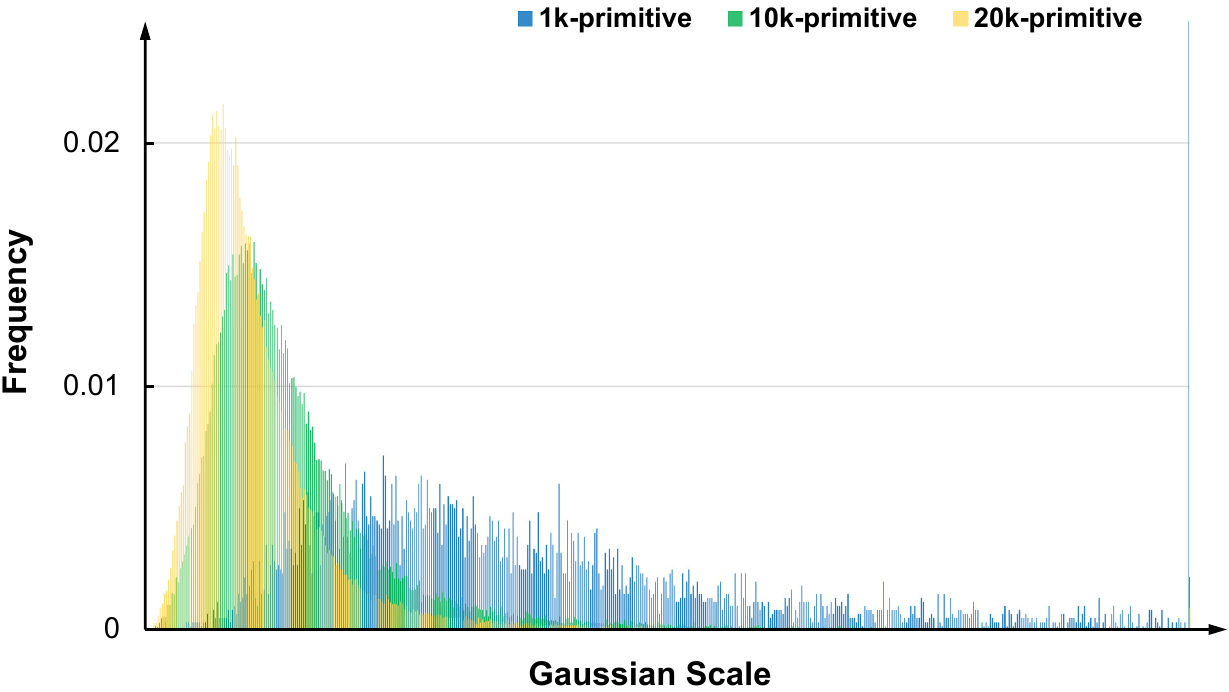}
    \caption{\textbf{The scale distribution of the Gaussians learned by models with different complexities.} The models with 10k and 20k primitives have a large portion of small-scale Gaussians. In contrast, the model with 1k primitives obtains more Gaussians with larger scales.}
    \label{img:scaling_distribution}
\end{figure}

On top of the above experiments, we further conduct experiments to study the Gaussian primitives in the trained models. Specifically, we visualize the scales of the Gaussians across different models in Fig.~\ref{img:scaling_distribution}. 
As we can see, the higher the model complexity is, the smaller the learned Gaussians are. Under this condition, the model is prone to overfitting the details in the training views without understanding the 3D structure. In contrast, the model with fewer Gaussian is encouraged to enlarge the Gaussian primitives to cover the contents in the training views. As a result, the 3D structure can be better modeled but high-frequency details cannot be well captured.

From the observations above, we can draw two challenges that limit the performance of 3DGS under sparse training views. \textit{First}, over-parameterized models suffer from the overfitting issue and produce degraded performance. \textit{Second}, decreasing the number of Gaussians alleviates the overfitting problem at the cost of the inferior capability to capture high-frequency details. To address these two challenges, we propose a coarse-to-fine framework with a random dropout regularization (RDR) to alleviate overfitting and an edge-guided splitting strategy (ESS) to remedy the missing details.

\subsection{Random Dropout Regularization (RDR)}
\label{sec:RDR}
\begin{figure}[t]
    \centering
	\begin{minipage}{0.49\linewidth}
		\centering
        \subfloat[gradients from $\mathcal{L}_{RDR}$]{\includegraphics[width=0.8\linewidth]{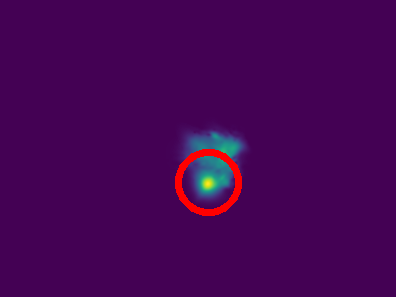} \label{img:analysis_grad_1}}
	\end{minipage}
	\begin{minipage}{0.49\linewidth}
		\centering
		\subfloat[gradients from $\mathcal{L}_{gt}$]{\includegraphics[width=0.8\linewidth]{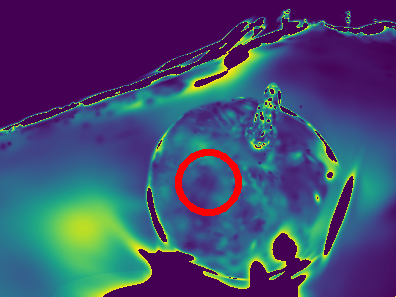} \label{img:analysis_grad_2}}
	\end{minipage}
    \caption{\textbf{Visualization of the gradient maps.} The dropouted Gaussian is annotated with a red circle. Thus brighter regions correspond to higher gradients.}
\end{figure}

During the training phase, the Gaussian primitives are randomly deactivated with a probability $p$ and the remaining Gaussian primitives are optimized to fit the observed views. Mathematically, the rendered color $\hat{C}$ for a pixel can be obtained as:
\begin{equation}
\left\{
\begin{aligned}
    &z \sim \text{Uniform}\left(0,1\right)\\
    &r = \left[z > p\right]\\
    &\hat{C} = \sum_{i\in \mathcal{N}}r_i\cdot \alpha_i \prod_{j=1}^{i-1}\left(1-r_j\cdot\alpha_j\right)c_i
\end{aligned}\right.,
\end{equation}
where $\text{Uniform}\left(0,1\right)$ denotes a uniform distribution, and $\left[\cdot\right]$ refers to Iverson bracket. During inference, all Gaussian primitives are activated for novel view synthesis. Particularly, we minimize the differences between the rendered results of the full model and the sub-model after dropout in each training iteration for optimization:
\begin{equation}
    \mathcal{L}_{RDR} = \left\|C-\hat{C}\right\|_1 + \text{SSIM}(C,\hat{C}).
    \label{eq:loss_function}
\end{equation}
Note that, instead of using the groundtruth color in the observed image for supervision, 
the rendered color $C$ produced by the full model is employed. To demonstrate the effectiveness of the loss function above, we splat the gradients of Gaussians to obtain gradient maps for comparison in Fig.~\ref{img:analysis_grad_1} and \ref{img:analysis_grad_2}.

If the rendered image of the full model is adopted as supervision, only neighboring Gaussian primitives centered at the dropouted one are optimized. As a result, the gradients only exist locally (Fig.~\ref{img:analysis_grad_1}) and encourage the model to focus on specific regions. In contrast, when the groundtruth color is employed for supervision, all Gaussian primitives are optimized and the gradients can be observed in all regions (Fig.~\ref{img:analysis_grad_2}). As a result, the gradients in different parts of a 3DGS model may counteract the effects of each other \cite{AbsGS}, producing inferior results. 

Although inspired by dropout, our method excludes its compensation strategy. Experiments show that the compensation strategy does not significantly enhance model performance and may alter the blending colors of the pixels unaffected by dropout, introducing potential adverse effects.

\noindent\textbf{Discussion.} 
Previous studies \cite{hara2016analysis, warde2013empirical, baldi2013understanding} have revealed that dropout mitigates overfitting by approximating the geometric mean over an ensemble of potential sub-networks. From this point of view, the optimization process of our RDR can be considered as training of several low-complexity sub-models. Then, the inference process can be viewed as an ensemble of the trained sub-models such that superior performance can be achieved. As shown in Fig.~\ref{img:motivation}, by integrating more low-complexity models, the ensemble model (\emph{i.e.}, the full model) produces consistent accuracy gains.

\begin{figure}[t]
    \centering
    \includegraphics[width=\linewidth]{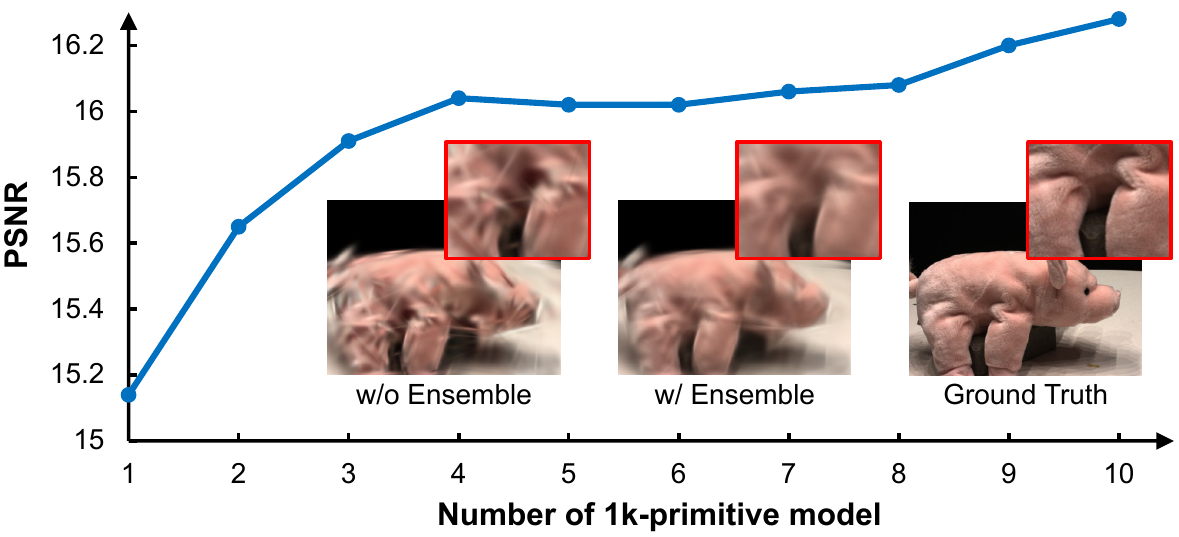}
    \caption{\textbf{Explaining the effectiveness of dropout from an ensemble learning perspective.} The quality of the rendered image significantly improves after integration, as evidenced by smoother appearances and fewer artifacts. This effect becomes more pronounced as the number of integrated sub-models increases.}
    \label{img:ensemble}
\end{figure}

\begin{figure*}[t]
    \centering
    \includegraphics[width=\linewidth]{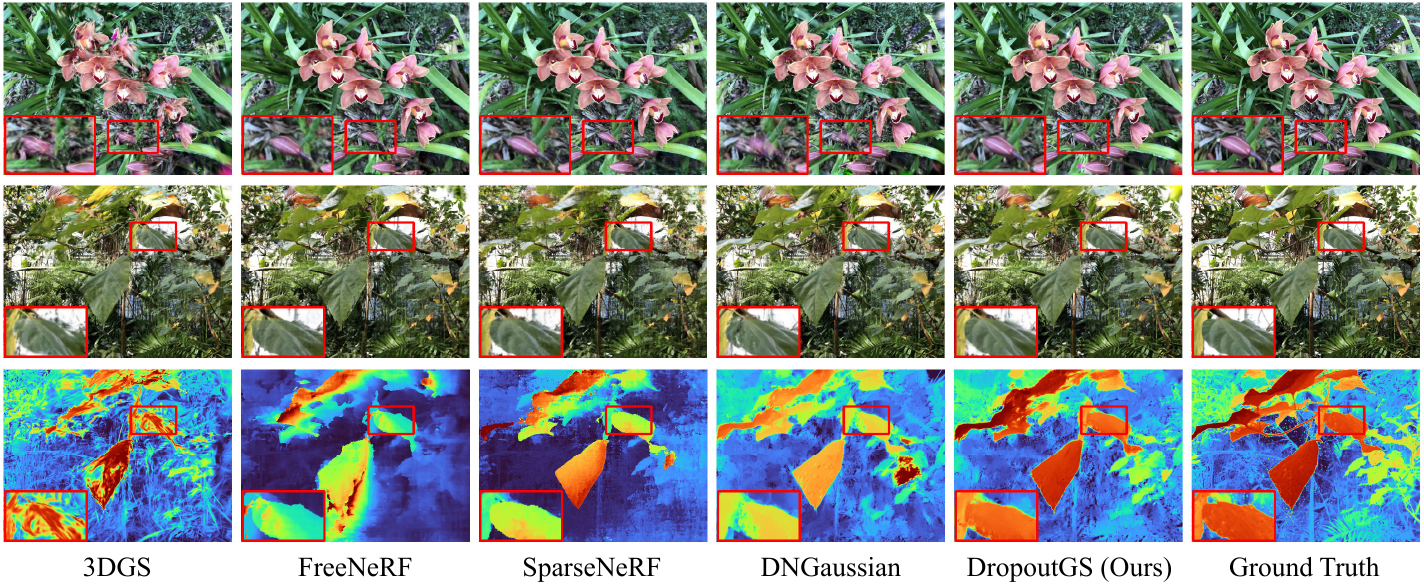}
    \caption{\textbf{Visual results on the LLFF dataset.} The ground truth of depth is obtained from the 3DGS trained with dense views. 3DGS fails to render novel views with correct geometry. DNGaussian can better represent the scene structure in comparison, but it cannot get smooth surfaces. The rendering results of FreeNeRF and SparseNeRF have a loss of details. Our model can learn complex details while obtaining smooth surfaces.}
    \label{img:llff}
\end{figure*}

\subsection{Edge-guided Splitting Strategy (ESS)}
\label{sec:ESS}
Although RDR facilitates 3DGS to better capture the 3D structure of the scene, the optimized Gaussian primitives are prone to being enlarged with a weakened capability to model high-frequency details. To remedy this, we propose an edge-guided splitting strategy (ESS) to further refine the Gaussian primitives by splitting the large ones into smaller ones to better fit the high-frequency details. 

\noindent\textbf{Edge Score.}
We first define an edge score to identify the edge regions. For an input view $I$, an edge detection method is employed to obtain pixel-wise probability $E(I)$, ranging from 0 to 1. Then, we project Gaussians onto the edge map and compute the single-view edge score $\mathcal{E}'$ by aggregating the probabilities of each primitive according to its contribution to pixels:
\begin{equation}
    \mathcal{E}_i' = \alpha_i\prod_j^{i-1}(1-\alpha_j)\cdot\sum_pE(I)\mathcal{M}^i(p)
\end{equation}
where $\mathcal{M}^{i}(p)$ is a binary mask indicating whether the \textit{p}-th pixel is covered by the \textit{i}-th Gaussian. Considering Gaussians cover different numbers of pixels in multiple viewpoints, we adopt a viewpoint accumulation approach to obtain the final edge score $\mathcal{E}_i$:
\begin{equation}
    \mathcal{E}_i = \sum_{k=1}^M\frac{\mathcal{E}_{i,k}'}{\sum_p\mathcal{M}_k^i}
\end{equation}
where $\mathcal{E}_{i,k}'$ indicates the edge score at the \textit{k}-th viewpoint out of a total of $M$. 

\noindent\textbf{Splitting Strategy.}
After obtaining the edge scores, Gaussian primitives with large scales and high edge scores are split into smaller ones to better fit the edge details. Specifically, the mask for the target primitives is calculated as below: 
\begin{equation}
    \mathcal{M}_{edge} = \left\{S_i \geq \mathcal{S}_{thr}\right\} \cap \left\{\mathcal{E}_i \geq \mathcal{E}_{thr}\right\},
\end{equation}
where $S_i$ denotes the size of \textit{i}-th primitives while $\mathcal{S}_{thr}$ and $\mathcal{E}_{thr}$ respectively denote the size threshold and edge threshold. By splitting large Gaussian primitives into smaller ones, finer high-frequency details can be modeled, as illustrated in Fig.~\ref{img:coarse-to-fine}.

\subsection{Loss Function}
\label{sec:loss}
Since DropoutGS only makes slight modifications to the rendering process and does not rely on additional modules, it can be seamlessly integrated with existing 3DGS techniques. In particular, our approach remains effective and provides further improvements when applied to methods with depth regularization. In this case, the overall loss function of DropoutGS is expressed as:
\begin{equation}
    \mathcal{L} = \mathcal{L}_{gs} + \lambda_{depth}\mathcal{L}_{depth} + \lambda_{RDR}\mathcal{L}_{RDR},
\end{equation}
where $\mathcal{L}_{gs}$ denotes the loss function in original 3DGS \cite{3DGS}. $\mathcal{L}_{depth}$ is the depth regularization used in DNGaussian and $\mathcal{L}_{RDR}$ denotes the proposed RDR loss. $\lambda_{depth}$ and $\lambda_{RDR}$ are the coefficients used to balance different constraints.

\begin{table*}[t]
\setlength{\abovecaptionskip}{4pt}
\resizebox{1\linewidth}{!}{
\setlength{\tabcolsep}{3.2 mm}
\begin{tabular}{l|c|cccccccc}
\toprule & & \multicolumn{4}{c}{LLFF} & \multicolumn{4}{c}{DTU} \\
 & \multirow{-2}{*}{Setting} & PSNR $\uparrow$ & LPIPS $\downarrow$ & SSIM $\uparrow$ & \multicolumn{1}{c|}{AVGE $\downarrow$} & PSNR $\uparrow$ & LPIPS $\downarrow$ & SSIM $\uparrow$ & AVGE $\downarrow$ \\ \midrule
SRF \cite{SRF} & & 12.34 & 0.591 & 0.250 & \multicolumn{1}{c|}{0.313} & 15.32 & 0.304 & 0.671 & 0.171 \\
PixelNeRF \cite{PixelNeRF}  & & 7.93 & 0.682 & 0.272 & \multicolumn{1}{c|}{0.461} & 16.82 & 0.270 & 0.695 & 0.147 \\

MVSNeRF \cite{MvsNeRF} & \multirow{-3}{*}{Trained on DTU} & 17.25 & 0.356 & 0.557 & \multicolumn{1}{c|}{0.171} & 18.63 & 0.197 &  0.769 & 0.113 \\ \midrule

SRF ft \cite{SRF}  & & 17.07 & 0.529 & 0.436 & \multicolumn{1}{c|}{0.203} & 15.68 & 0.281 & 0.698 & 0.162 \\

PixelNeRF ft \cite{PixelNeRF} & & 16.17 & 0.512 & 0.438 & \multicolumn{1}{c|}{0.217} & 18.95 & 0.269 & 0.710 & 0.125 \\

MVSNeRF ft \cite{MvsNeRF} & \multirow{-3}{*}{\begin{tabular}[c]{@{}c@{}}Trained on DTU \\ Fine-tuned per Scene\end{tabular}} & 17.88 &  0.327 & 0.584 & \multicolumn{1}{c|}{0.157} & 18.54 & 0.197 &  0.769 & 0.113 \\ \midrule

Mip-NeRF \cite{MipNeRF} & & 14.62 & 0.495 & 0.351 & \multicolumn{1}{c|}{0.246} & 8.68  & 0.353 & 0.571 & 0.323 \\

DietNeRF \cite{DietNeRF}  & & 14.94 & 0.496 & 0.370 & \multicolumn{1}{c|}{0.240} & 11.85 & 0.314 & 0.633 & 0.243 \\

RegNeRF \cite{RegNeRF} & & 19.08 & 0.336 & 0.587 & \multicolumn{1}{c|}{0.149} & 18.89 &  0.190 & 0.745 & 0.112 \\

FreeNeRF \cite{FreeNeRF} & & \cellcolor[HTML]{FFE4CF}\underline{19.63} & \cellcolor[HTML]{FFFFD4}0.308 & \cellcolor[HTML]{FFFFD4}0.612 & \multicolumn{1}{c|}{ 0.134} & \cellcolor[HTML]{FFE4CF}\underline{19.92} & \cellcolor[HTML]{FFFFD4}0.182 & \cellcolor[HTML]{FFFFD4}0.787 & \cellcolor[HTML]{FFE4CF}\underline{0.098} \\

SparseNeRF \cite{SparseNeRF} & & \cellcolor[HTML]{FFCCC9}\textbf{19.86} & 0.328 & \cellcolor[HTML]{FFCCC9}\textbf{0.624} & \multicolumn{1}{c|}{\cellcolor[HTML]{FFCCC9}\textbf{0.127}} & \cellcolor[HTML]{FFFFD4}19.55 & 0.201 &  0.769 & \cellcolor[HTML]{FFFFD4}0.102 \\ 

3DGS \cite{3DGS} & &  15.52 & 0.405 & 0.408 & \multicolumn{1}{c|}{0.209} &  10.99 & 0.313 & 0.585 & 0.252 \\

3DGS\dag & & 16.46 & 0.401 & 0.440 & \multicolumn{1}{c|}{0.192} & 14.74 &  0.249 &  0.672 & 0.169 \\

DNGaussian \cite{DNGS} &  & 19.12 & \cellcolor[HTML]{FFE4CF}\underline{0.294} & 0.591 & \multicolumn{1}{c|}{\cellcolor[HTML]{FFFFD4}0.132} & 18.91 & \cellcolor[HTML]{FFE4CF}\underline{0.176} & \cellcolor[HTML]{FFE4CF}\underline{0.790} & \cellcolor[HTML]{FFFFD4}0.102 \\

\textbf{DropoutGS (Ours)} & \multirow{-9}{*}{Optimized per Scene}& \cellcolor[HTML]{FFFFD4}19.35 & \cellcolor[HTML]{FFCCC9}\textbf{0.282} & \cellcolor[HTML]{FFE4CF}\underline{0.622} & \multicolumn{1}{c|}{\cellcolor[HTML]{FFE4CF}\underline{0.128}} & \cellcolor[HTML]{FFCCC9}\textbf{20.22} & \cellcolor[HTML]{FFCCC9}\textbf{0.150} & \cellcolor[HTML]{FFCCC9}\textbf{0.830} & \cellcolor[HTML]{FFCCC9}\textbf{0.086} \\  
\bottomrule
\multicolumn{10}{l}{\small\dag \ with the same hyperparameters and the neural color renderer as DNGaussian.}
\end{tabular}
}
\caption{\textbf{Quantitative comparison on the LLFF and DTU datasets.} The best, second-best, and third-best entries are marked in \colorbox[HTML]{FFCCC9}{\textbf{red}}, \colorbox[HTML]{FFE4CF}{\underline{orange}}, and \colorbox[HTML]{FFFFD4}{yellow}, respectively. }
\label{tab:llffdtu}
\vspace{-.4cm}
\end{table*}

\begin{table}[t]
\setlength{\abovecaptionskip}{4pt}
\resizebox{1\linewidth}{!}{
\setlength{\tabcolsep}{5 mm}
\centering
\begin{tabular}{l|ccc}
\toprule
Method & PSNR $\uparrow$ & SSIM $\uparrow$ & LPIPS $\downarrow$ \\ \midrule
NeRF  & 14.934 & 0.687 & 0.318  \\
NeRF (Simplified)   & 20.092  & 0.822 & 0.179 \\
DietNeRF    & 23.147 & 0.866 & 0.109 \\
DietNeRF + ft  & 23.591& 0.874 & \cellcolor[HTML]{FFFFD4}0.097 \\
FreeNeRF  & \cellcolor[HTML]{FFFFD4}24.259& \cellcolor[HTML]{FFFFD4}0.883 & 0.098 \\ 
SparseNeRF  & 22.410 & 0.861 & 0.119 \\ \midrule
3DGS  & 22.226 & 0.858 & 0.114 \\
DNGaussian& \cellcolor[HTML]{FFE4CF}\underline{24.305}& \cellcolor[HTML]{FFE4CF}\underline{0.886}& \cellcolor[HTML]{FFE4CF}\underline{0.088} \\
\textbf{DropoutGS (Ours)}& \cellcolor[HTML]{FFCCC9}\textbf{24.476}& \cellcolor[HTML]{FFCCC9}\textbf{0.889}& \cellcolor[HTML]{FFCCC9}\textbf{0.085}\\ 
\bottomrule
\end{tabular}
}
\caption{\textbf{Quantitative Comparison on Blender for 8 input views.} The best, second-best, and third-best entries are marked in \colorbox[HTML]{FFCCC9}{\textbf{red}}, \colorbox[HTML]{FFE4CF}{\underline{orange}}, and \colorbox[HTML]{FFFFD4}{yellow}, respectively.}
\label{tab:blender}
\vspace{-.4cm}
\end{table}
\section{Experiments}
\subsection{Setups}
\noindent\textbf{Datasets.} 
We evaluate our method on three benchmark datasets: the LLFF \cite{LLFF} dataset, the DTU \cite{DTU} dataset and the Blender \cite{NeRF} dataset. 
We follow the 3-view setting used in previous works \cite{DNGS, FreeNeRF, SparseNeRF} with the same split of LLFF and DTU. 
Additionally, we adopt the same object masks as \cite{FreeNeRF} for fair comparison on the rendering quality of the target objects rather than the background in DTU. For Blender, we train our model with 8 views and evaluate it on the other 25 views following the setting used in \cite{DietNeRF, FreeNeRF}.

\noindent\textbf{Evaluation Metrics.} 
We report PSNR, SSIM \cite{SSIM} and LPIPS \cite{LPIPS} metrics to evaluate the rendering quality. Besides, we also adopt the Average Error (AVGE) \cite{MipNeRF} as a metric, which is the geometric mean of ${\rm MSE}$, $\sqrt{1-{\rm{SSIM}}}$ and ${\rm LPIPS}$. All results are averaged over three repeated experiments.

\noindent\textbf{Baselines.} 
We compare our DropoutGS with 9 state-of-the-art methods, including one 3DGS-based method (DNGaussian \cite{DNGS}), 5 NeRF-based methods (FreeNeRF \cite{FreeNeRF}, SparseNeRF \cite{SparseNeRF}, RegNeRF \cite{RegNeRF}, DietNeRF \cite{DietNeRF}, and Mip-NeRF \cite{MipNeRF}) and 3 generative methods (MVSNeRF \cite{MvsNeRF}, PixelNeRF \cite{PixelNeRF}, SRF \cite{SRF}). 

\noindent\textbf{Implementations.} 
Our approach is built on the PyTorch framework. We train our model for 6k iterations on the LLFF, DTU, and Blender datasets. Following DNGaussian \cite{DNGS}, we initialize 3DGS and DropoutGS with a randomly generated point cloud.

\subsection{Performance Evaluation}
\noindent\textbf{The LLFF Dataset.} 
We evaluate our method on eight complex scenes of the LLFF dataset with 3 views as input. Quantitative results are presented in Table~\ref{tab:llffdtu} while visual results are shown in Fig.~\ref{img:llff}.

It can be observed from Table~\ref{tab:llffdtu} that our method outperforms other methods in terms of LPIPS and achieves the second-best performance in SSIM and AVGE. This demonstrates the superior quantitative performance of our method. Moreover, as shown in Fig.~\ref{img:llff}, our DropoutGS produces results with finer details and fewer artifacts, while FreeNeRF \cite{FreeNeRF} and SparseNeRF \cite{SparseNeRF} generate blurry results with inferior perceptual quality. Although DNGaussian also captures fine details, it suffers from severe hollow artifacts. Benefiting from our RDR and ESS strategies, our DropoutGS fills up the hollows caused by overfitting without sacrificing details. Additionally, we further visualize the depth map produced by different methods. As compared to DNGaussian, our DropouGS generates more accurate depth maps with smooth surfaces, which means the 3D structure is better modeled.

\begin{figure}[t]
    \centering
	\begin{minipage}{\linewidth}
		\centering
        \subfloat[Rendered images visualization]{\includegraphics[width=\linewidth]{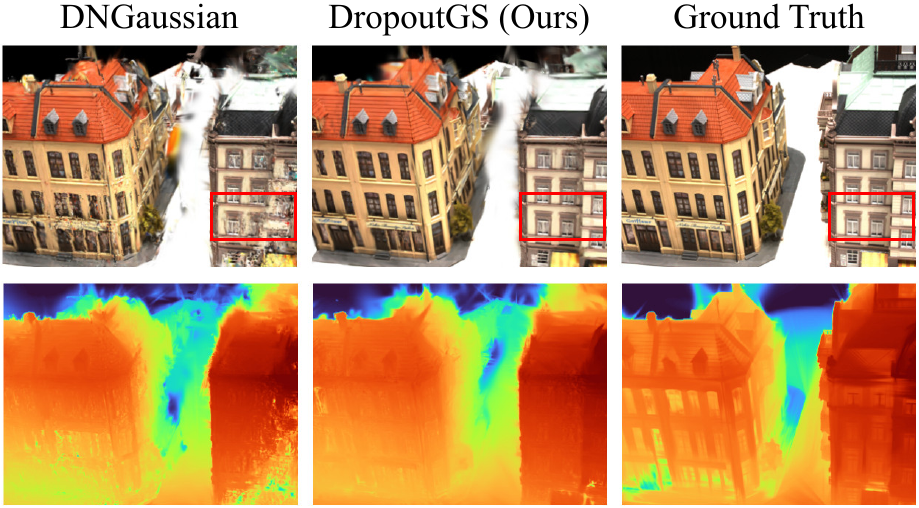}}
	\end{minipage}\\
	\begin{minipage}{\linewidth}
		\centering
		\subfloat[Gaussian point cloud visualization]{\includegraphics[width=\linewidth]{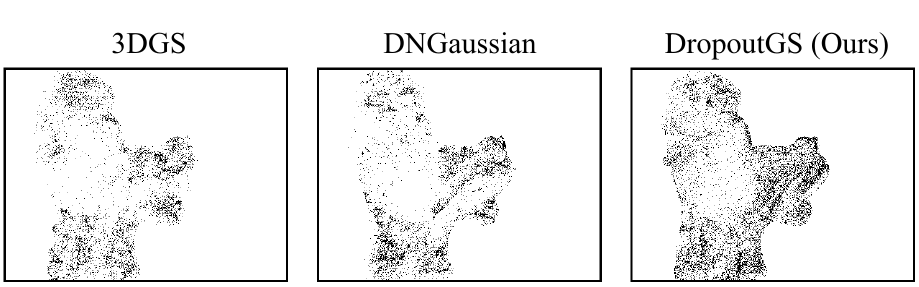}}
	\end{minipage}
    \caption{\textbf{Qualitative results on the DTU dataset.} The ground truth of depth is obtained from the 3DGS trained with dense views. Our method learns a denser primitive distribution improving the geometry accuracy.}
    \label{img:dtu}
\end{figure}

\noindent\textbf{The DTU Dataset.} 
To evaluate the rendering ability of our method on the object-centered scenes, we conduct our experiments on the DTU dataset. As shown in Table~\ref{tab:llffdtu}, our proposed method significantly outperforms all the baselines across multiple metrics in the 3-view setting, achieving state-of-the-art performance. Figure~\ref{img:dtu} illustrates that DropoutGS successfully renders novel views with accurate geometric structure and reduced overfitting artifacts such as hollows and blurring, while DNGaussian suffers from severe degradation in the foreground. Additionally, our DropoutGS produces denser point clouds that are better aligned with the surfaces of objects. This further demonstrates the effectiveness of our DropoutGS.

\begin{figure}[t]
    \centering
    \includegraphics[width=1\linewidth]{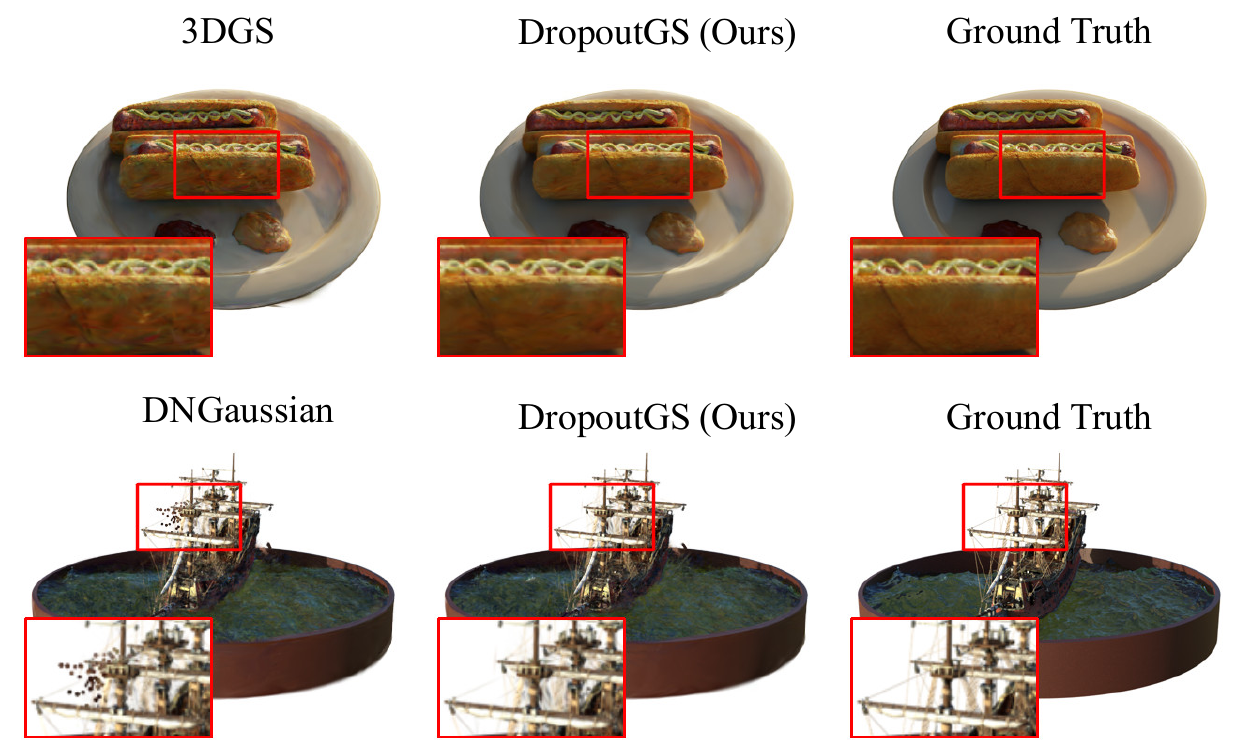}
    \caption{\textbf{Qualitative results on the Blender dataset.} Our DropoutGS can smooth out the artifacts that appear in 3DGS and DNGaussian, obtaining cleaner and smoother novel views.}
    \label{img:blender}
    \vspace{-.3cm}
\end{figure}

\noindent\textbf{The Blender Dataset.}
We also evaluate our model on the Blender dataset with 8-view input, which contains eight synthetic scenes each with widely varying training and testing viewpoints. The quantitative results in Table~\ref{tab:blender} show that the proposed method achieves the best performance in terms of PSNR, SSIM, and LPIPS. From the qualitative results in Fig.~\ref{img:blender}, we can see that DropoutGS is able to generate views with fewer floaters compared to the vanilla 3DGS and the original DNGaussian. The $\textit{hotdog}$ scene also demonstrates the ability of our method to smooth out inadequately trained Gaussian primitives, resulting in a flatter surface. These experiments show that DropoutGS achieves competitive performance not only on the forward-facing dataset such as LLFF and DTU but also on the wide-baseline dataset with challenging reflective materials.

\noindent\textbf{Compatibility.} 
Our DropoutGS is compatible with different 3DGS methods and can be flexibly integrated with them to improve their performance under sparse views. To demonstrate this, we combine our DropoutGS with the original 3DGS, FSGS \cite{FSGS}, and CoR-GS \cite{CoRGS} for evaluation. As listed in Table~\ref{tab:generalization}, our approach facilitates these methods to produce notable gains in terms of all metrics. Unlike FSGS and CoR-GS which employ the point cloud obtained from \cite{sfm} for initialization, Gaussians in 3DGS are randomly initialized so that they are more susceptible to the overfitting issue. As a result, our method introduces more significant improvements in 3DGS. Remarkably, our method is competitive with different initialization methods and produces consistent performance gains.

\begin{table}[t]
\setlength{\abovecaptionskip}{4pt}
\resizebox{1\linewidth}{!}{
\setlength{\tabcolsep}{3 mm}
\begin{tabular}{l|c|c|cccc}
\toprule
Method & w/ Ours & Setting & PSNR $\uparrow$  & LPIPS $\downarrow$ & SSIM $\uparrow$ & AVGE $\downarrow$  \\ \midrule
& \ding{55}&& 16.46 & 0.401 & 0.440 & 0.192 \\
\multirow{-2}{*}{3DGS\dag}& \checkmark &\multirow{-2}{*}{Random init.}&\textbf{18.05}  &\textbf{0.326} & \textbf{0.545} & \textbf{0.155} \\\midrule

&\ding{55} &&19.86&0.222&0.670&0.112\\
\multirow{-2}{*}{3DGS\dag *}&\checkmark&\multirow{-2}{*}{MVS init.}&\textbf{20.53}&\textbf{0.205}&\textbf{0.706}&\textbf{0.102}\\\midrule

&\ding{55} && 20.45 & 0.196 & 0.712  & 0.101\\
\multirow{-2}{*}{CoR-GS* \cite{CoRGS}}&\checkmark &\multirow{-2}{*}{Co-constraint}& \textbf{20.59} & \textbf{0.195}  & \textbf{0.716} & \textbf{0.097}\\\midrule

&\ding{55} && 20.43& 0.248& 0.682& 0.108\\
\multirow{-2}{*}{FSGS* \cite{FSGS}}&\checkmark & \multirow{-2}{*}{Depth constraint}& \textbf{20.70}&\textbf{0.200} &\textbf{0.713} &\textbf{0.099}\\\midrule

&\ding{55} && 19.94& 0.228& 0.682& 0.116\\
\multirow{-2}{*}{DNGaussian*}& \checkmark &\multirow{-2}{*}{Depth constraint}&\textbf{20.64}  &\textbf{0.210} & \textbf{0.717} & \textbf{0.101} \\
\bottomrule
\end{tabular}
}
\caption{\textbf{Compatibility to different 3DGS methods with 3 input views on the LLFF dataset.} \textit{init.} denotes initialization. * with the MVS point cloud as initialization.}
\label{tab:generalization}
\vspace{-.4cm}
\end{table}

\subsection{Ablation Study}
\label{sec:ablation}
\begin{figure}[ht]
    \centering
    \includegraphics[width=\linewidth]{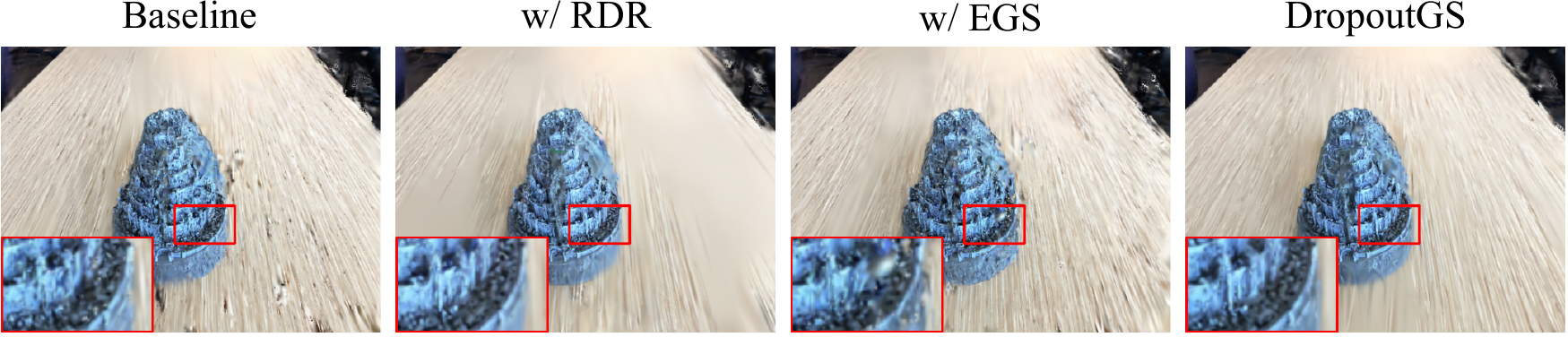}
    \caption{\textbf{Qualitative results of ablation study.} The model with only ESS can cause over-splitting, where too many insufficiently trained Gaussians are generated in high-frequency regions as artifacts. Only the full model can generate clean and sharp novel views.}
    \label{img:ablation}
\end{figure}

In this subsection, we conduct ablation experiments to study the effects of our DropoutGS. All experiments are conducted on the LLFF dataset with 3 views input. Quantitative and qualitative results are reported in Table~\ref{tab:ablation} and Fig.~\ref{img:ablation}.

\noindent\textbf{RDR and ESS Strategies.} 
We first conduct experiments to investigate the effectiveness of RDR and ESS strategies. Specifically, we introduce these two strategies to the baseline method separately and compare their performance with our DropoutGS. It can be observed from Table~\ref{tab:ablation} that both RDR and ESS introduce notable performance improvement to the baseline. It is worth noticing that RDR produces a significant gain on PSNR and SSIM, but introduces a slight drop in LPIPS. With only RDR, Gaussian primitives are prone to be large such that high-frequency details cannot be well reconstructed. With both additional ESS, the large Gaussian primitives are split into small ones to better capture details in edge regions. As a result, the LPIPS score is improved together with PSNR and SSIM. Figure~\ref{img:ablation} further compares the visual results produced by methods with different settings. {It can be seen that RDR helps to fill in the hollows caused by overfitting and ESS contributes to recovering the high-frequency details in the rendered views.}

\begin{table}[t]
\setlength{\abovecaptionskip}{5pt}
\setlength{\tabcolsep}{3.5 mm}
\resizebox{1\linewidth}{!}{
    \centering
    \begin{tabular}{l|cc|cccc}
    \toprule
    &RDR&  ESS&  PSNR$\uparrow$&  SSIM$\uparrow$& LPIPS$\downarrow$ &AVGE$\downarrow$\\\midrule
    Baseline &&  &  18.76&  0.582&  0.300&0.139\\
    w/ ESS&&  $\checkmark $&  18.91&  0.592&  0.294&0.136\\
    w/ RDR& $\checkmark$ & & 19.26&0.608&0.317&0.134\\
    DropoutGS& $\checkmark $& $\checkmark $& \textbf{19.35}& \textbf{0.622}& \textbf{0.282}&\textbf{0.128}\\\bottomrule
    \end{tabular}}
    \caption{\textbf{Quantitative results of ablation study.} We ablate our model with 3-view input on the LLFF dataset. ``Baseline" denotes DNGaussian.}
    \label{tab:ablation}
\end{table}

\noindent\textbf{More View Settings.}
In addition to the 3-view input, we further evaluate the performance of DropoutGS in a more view setting on the LLFF dataset. We respectively use 6 views and 9 views as input following the default settings in DNGaussian and report the results in Table~\ref{tab:more_views}. Experiments show that the DropoutGS achieves better performance as the training view increases and outperforms the baselines at all training settings. However, it can be observed that the performance improvement of DropoutGS decreases at denser viewpoint settings. This is mainly because denser viewpoints provide more constraints to the model during training, thus reducing its dependence on external constraints.

\begin{table}[t]
\setlength{\abovecaptionskip}{4pt}
\resizebox{1\linewidth}{!}{
\setlength{\tabcolsep}{3 mm}
\begin{tabular}{c|l|cccc}
\toprule
\multicolumn{1}{l|}{Views} & Method     & PSNR $\uparrow$  & LPIPS $\downarrow$ & SSIM $\uparrow$ & AVGE $\downarrow$  \\ \midrule
\multirow{4}{*}{3}& 3DGS       & 15.52 & 0.405 & 0.408 & 0.209 \\
                           & 3DGS\dag   & 16.46 & 0.401 & 0.440 & 0.192 \\
                           & DNGaussian & 19.12& 0.294& 0.591& 0.132\\
 & \textbf{DropoutGS (Ours)}& \textbf{19.23}& \textbf{0.287}& \textbf{0.614}&\textbf{0.130}\\ \midrule
\multirow{4}{*}{6}& 3DGS       & 20.63 & 0.226 & 0.699 & 0.108 \\
                           & 3DGS\dag   & 21.09 & 0.229 & 0.699 & 0.103 \\
                           & DNGaussian & 22.18& 0.198& 0.755& 0.088\\
 & \textbf{DropoutGS (Ours)}& \textbf{23.35}& \textbf{0.177}& \textbf{0.791}&\textbf{0.076}\\ \midrule
\multirow{4}{*}{9}& 3DGS       & 20.44 & 0.230 & 0.697 & 0.108 \\
                           & 3DGS\dag   & 23.21& 0.176& 0.785 & 0.076\\
                           & DNGaussian & 23.17 & 0.180 & 0.788& 0.077 \\ 
 & \textbf{DropoutGS (Ours)}& \textbf{24.33}& \textbf{0.160}& \textbf{0.825}& \textbf{0.066}\\ \bottomrule
\end{tabular}
}
\caption{\textbf{Comparison with 3, 6, and 9 input views on LLFF.}}
\label{tab:more_views}
\vspace{-.3cm}
\end{table}

\begin{figure}[t]
    \centering
    \includegraphics[width=\linewidth]{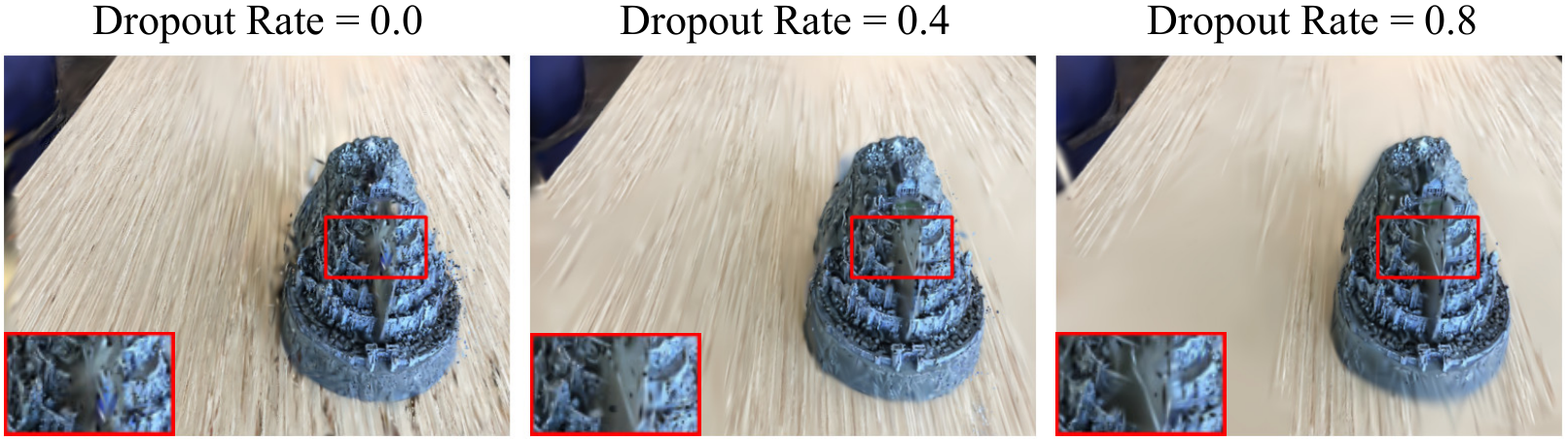}
    \caption{\textbf{Qualitative results of the ablation study on the random dropout rate.} A large dropout rate results in a reduction of artifacts but also a loss of details. }
    \label{img:dropoutrate}
\end{figure}

\noindent\textbf{Dropout Rate.}
{
We conduct ablation experiments on the dropout rate to investigate its impact on RDR performance. Table~\ref{tab:dropoutrate} reports the performance of DropoutGS with varying dropout rates using 3 views with visual results being shown in Fig.~\ref{img:dropoutrate}. As the dropout rate increases from 0.0 to 0.2, quantitative performance gains are achieved in terms of all metrics with fewer visual artifacts. When the dropout rate continues to increase, although higher PSNR is produced, the LPIPS score suffers a slight degradation due to excessive smoothing of the resultant images.}

\begin{table}[t]
\setlength{\abovecaptionskip}{4pt}
\resizebox{1\linewidth}{!}{
\setlength{\tabcolsep}{3 mm}
\begin{tabular}{c|cccc}
\toprule
Dropout Rate & PSNR $\uparrow$  & SSIM $\uparrow$ & LPIPS $\downarrow$ & AVGE $\downarrow$  \\ \midrule
0.0& 18.76 & 0.582 & 0.300 & 0.139 \\
0.2& 19.20 & 0.616 & 0.279 & 0.130 \\
0.4& 19.35 & 0.622 & 0.282 & 0.128 \\ 
0.6& 19.37 & 0.622 & 0.298 & 0.130 \\
0.8& 19.44 & 0.621 & 0.302 & 0.129 \\
\bottomrule
\end{tabular}
}
\caption{\textbf{Ablation results on the random dropout rate.} We conduct the 
reproduced performance of DNGaussian as the results without random dropout regularization.}
\label{tab:dropoutrate}
\end{table}

\begin{figure}[t]
    \centering
    \includegraphics[width=\linewidth]{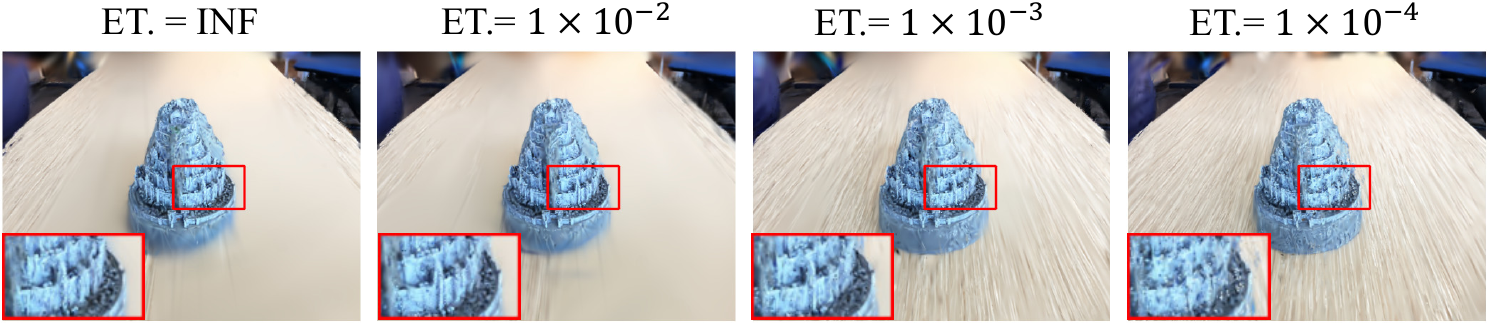}
    \caption{\textbf{The visualization result of the ablation study on the edge threshold.} The highlighted foreground region appears overfitting degradation at excessively small edge thresholds.}
    \label{img:edgethreshold}
    \vspace{-.3cm}
\end{figure}

\noindent\textbf{Edge Threshold.} { We further conduct experiments to investigate the impact of the edge threshold in our ESS strategy. Quantitative and qualitative results are presented in Table~\ref{tab:edgethreshold} and Fig.~\ref{img:edgethreshold}. As the edge threshold decreases, high-frequency details can be better preserved such that the LPIPS score is consistently improved from 0.317 to 0.275. However, excessively low thresholds lead to notable distortion with degraded PSNR performance. Overall, 0.001 is selected as the default setting of the edge threshold.
}

\begin{table}[t]
\setlength{\abovecaptionskip}{4pt}
\resizebox{1\linewidth}{!}{
\setlength{\tabcolsep}{3 mm}
\begin{tabular}{c|cccc}
\toprule
Edge Threshold (ET.) & PSNR $\uparrow$  & SSIM $\uparrow$ & LPIPS $\downarrow$ & AVGE $\downarrow$  \\ \midrule
INF& 19.26&0.608&0.317&0.134\\
$1\times 10^{-1}$& 19.22&0.608&0.316&0.134\\
$1\times 10^{-2}$& 19.29&0.611&0.313&0.133\\
$1\times 10^{-3}$& 19.35 & 0.622 & 0.282 & 0.128 \\ 
$1\times 10^{-4}$& 19.17 & 0.618 & 0.275 & 0.129 \\
\bottomrule
\end{tabular}}
\caption{\textbf{Ablation results on the edge threshold.} We explore the performance of different edge thresholds on the LLFF dataset with 3 views as input. }
\label{tab:edgethreshold}
\end{table}

\section{Conclusion}
{In this paper, we present a novel coarse-to-fine framework, termed DropoutGS, for sparse view rendering. Specifically, we introduce the dropout technique into 3DGS and propose a random dropout regularization to alleviate overfitting. Then, an edge-guided splitting strategy is designed for further recovery of the high-frequency details. Extensive experiments validate the effectiveness and superiority of our approach over existing methods and demonstrate its compatibility with 3DGS-based techniques.}
\section*{Acknowledgements}
This work was partially supported by the National Natural Science Foundation of China (No.~U20A20185, 62372491, 62301601), the Guangdong Basic and Applied Basic Research Foundation (No.~2022B1515020103, 2023B1515120087), the Shenzhen Science and Technology Program (No.~RCYX20200714114641140), the Science and Technology Research Projects of the Education Office of Jilin Province (No.~JJKH20251951KJ), and the SYSU-Sendhui Joint Lab on Embodied AI.

{
    \small
    \bibliographystyle{ieeenat_fullname}
    \bibliography{main}
}


\end{document}